\documentclass[pre,aps,twocolumn]{revtex4}
\usepackage{amsmath}
\usepackage{amssymb}
\usepackage{graphicx}
\usepackage{subfigure}
\usepackage{algorithm}
\usepackage{algpseudocode}
\usepackage{bm}
\usepackage{color}
\usepackage{xcolor}

\usepackage{commath}
\usepackage{dcolumn}

\usepackage{hyperref} 

\newcommand{\cut}[1]{{}}

\begin{document}

\title{
  Active online learning in the binary perceptron problem
}

\author{Hai-Jun Zhou$^{1,2}$}

\affiliation{
  $^1$CAS Key Laboratory for Theoretical Physics, Institute of Theoretical Physics, Chinese Academy of Sciences, Beijing 100190, China\\
  $^2$School of Physical Sciences, University of Chinese Academy of Sciences, Beijing 100049, China
}
\date{\today}

\begin{abstract}
  The binary perceptron is the simplest artificial neural network formed by $N$ input units and one output unit, with the neural states and the synaptic weights all restricted to $\pm 1$ values. The task in the teacher--student scenario is to infer the hidden weight vector by training on a set of labeled patterns. Previous efforts on the passive learning mode have shown that learning from independent random patterns is quite inefficient. Here we consider the active online learning mode in which the student designs every new Ising training pattern. We demonstrate that it is mathematically possible to achieve perfect (error-free) inference using only $N$ designed training patterns, but this is computationally unfeasible for large systems.  We then investigate two Bayesian statistical designing protocols, which require $2.3 N$ and $1.9 N$ training patterns, respectively, to achieve error-free inference. If the training patterns are instead designed through deductive reasoning, perfect inference is achieved using $N\!+\!\log_{2}\!N$ samples. The performance gap between Bayesian and deductive designing strategies may be shortened in future work by taking into account the possibility of ergodicity breaking in the version space of the binary perceptron.
\end{abstract}

\maketitle

\section{Introduction}

The perceptron invented by Frank Rosenblatt in 1957 is probably the simplest artificial neural network~\cite{Rosenblatt-1958}. It has $N$ inputs and one output, with each input neuron $i$ affecting the output neuron through a synapse of weight $T_i$~\cite{Watkin-etal-1993,Engel-VanDenBroeck-2001}.  Given an $N$-dimensional input vector $\bm{\xi}\equiv (\xi_1, \xi_2, \ldots, \xi_N)$, the binary output state $\sigma$ is determined according to a (highly nonlinear) sign function
\begin{equation}
  \label{eq:signf}
  \sigma(\bm{\xi}) = {\rm sign}\bigl(\bm{\xi}; \bm{T}\bigr)
  \equiv {\rm sign}\Bigl( \sum\limits_{i=1}^{N} T_i \xi_i \Bigr) \; ,
\end{equation}
where $\bm{T} \equiv (T_1, T_2, \ldots, T_N)$ denotes the synaptic weight vector. The output $\sigma=1$ if the overlap between $\bm{\xi}$ and $\bm{T}$, defined as $\sum_{i} T_i s_i$, is positive and $\sigma=-1$ if this overlap is negative. We consider the binary (or Ising) perceptron problem, so all the synaptic weights and neural states are restricted to be Ising-valued (i.e.,  $T_i, \xi_i \in  \pm 1$ for every input neuron $i$ of the system). These constraints make the binary perceptron much more challenging to study than the continuous counterpart~\cite{Watkin-etal-1993,Engel-VanDenBroeck-2001}. 
 
The perceptron can serve as a linear classifier. In this scenario, given $P$ patterns $\bm{\xi}^\mu \equiv (\xi_1^\mu, \xi_2^\mu, \ldots, \xi_N^\mu)$ with indices $\mu = 1, 2, \ldots, P$ and their binary labels $\sigma^\mu$, the task is to find a vector $\bm{T}$ such that these $P$ patterns are correctly classified, that is, $\sigma^\mu = {\rm sign}(\bm{\xi}^\mu; \bm{T})$ holds for each and every one of these $P$ patterns~\cite{Watkin-etal-1993,Engel-VanDenBroeck-2001}. The whole set of all such compatible weight vectors form the version space (or solution space) of the perceptron. If the input binary patterns $\bm{\xi}^\mu$ are random and independent, it was predicted using the replica method of statistical physics that there exist binary solutions $\bm{T}$ to this classification problem as long as $P < 0.83 N$~\cite{Krauth-Mezard-1989}. Several message-passing algorithms~\cite{Kabashima-Uda-2004,Braunstein-Zecchina-2006} inspired by the cavity method of statistical physics have been implemented to solve single instances of the perceptral classification problem. More recently it was revealed that, as the density $\alpha \equiv P/N$ of random input binary patterns increases, the typical (equilibrium) solutions  of this classification problem become widely separated from each other and are extremely hard to reach~\cite{Huang-Kabashima-2014,Huang-etal-2013}, while there are also sub-dominant dense regions in the version space which could be reached through entropy-weighted sampling strategies~\cite{Baldassi-etal-2015,Obuchi-Kabashima-2009}.

The perceptron can also be studied from the teacher--student perspective, with $\bm{T}$ understood as the teacher's weight vector which is hidden to the student. For each query $\bm{\xi}$ to the teacher, the correct label $\sigma(\bm{\xi})$ as computed through Eq.~(\ref{eq:signf}) is revealed to the student, and the task for the student is to infer the hidden vector $\bm{T}$ by learning from the $(\bm{\xi}, \sigma)$ associations~\cite{Watkin-etal-1993,Engel-VanDenBroeck-2001}. This inference task may be carried out in an offline manner, meaning that the training samples are repeatedly examined by the student during the learning process. It may also be carried out in an online manner, meaning that each training sample is used only once by the student to update his/her belief on $\bm{T}$. We study online learning in the present work.

The learning performance of the teacher--student perceptron system has been investigated by many authors. In the passive learning mode for which the training patterns are independent and random, it was predicted that perfect (error-free) inference of a binary vector $\bm{T}$ is theoretically possible with $P \approx 1.25 N$ binary samples~\cite{Gyorgyi-1990,Sompolinsky-etal-1990}. But this theoretical limit has never been achieved by actual heuristic algorithms. Theoretical and numerical studies on various algorithms have found that the generalization error $\varepsilon$ of the passive learning process decreases with the pattern density $\alpha$ algebraically, e.g., $\varepsilon \propto \alpha^{-1}$. This means that in the thermodynamic limit of $N\rightarrow \infty$, perfect inference is unlikely to achieve at any finite value of pattern density $\alpha$~\cite{Littlestone-Warmuth-1989,Opper-Haussler-1991,Seung-Opper-etal-1992,Opper-1996,Feng-1998,RosenZvi-2000,Baldassi-2009}.

In this work we address the issue of active learning, which aims at accelerating online inference by carefully designing the training patterns. After the student has encountered $P$ samples and has already gained some knowledge about the truth vector $\bm{T}$, how should she/he design the $(P+1)$-th query so that the answer from this new query will be most informative for inference? This interesting question was explored by many authors in the early 1990s (see, e.g., Refs.~\cite{Kinzel-Rujan-1990,Baum-1991,Hwang-etal-1991,Kinouchi-Caticha-1992,Seung-etal-1992,Watkin-Rau-1992,Kabashima-Shinomoto-1993,Sollich-Saad-1995}), but the focus was on minimization of the generalization error rather than on error-free inference. In Section~\ref{sec:logic} of the present paper we prove that error-free learning of $\bm{T}$ can be achieved using at most $N\!+\!\log_{2}\!N$ designed training patterns through deductive reasoning, which is only slightly beyond the theoretical lower bound, $N$. If the optimal Bayesian inference strategy is used instead of deductive logic, we find that error-free inference using exactly $N$ designed samples is indeed possible, but it is computationally feasible only for small systems (Section~\ref{sec:enumerate}). We then implement two heuristic designing algorithms in Sections~\ref{sec:hs1} and \ref{sec:hs2} for large systems based on this optimal Bayesian principle. Our simulation results demonstrate that these two heuristic algorithms need $2.3 N$ and $1.9 N$ training samples, respectively, to achieve error-free inference.

Although the deductive-logic algorithm certainly outperforms the data-driven Bayesian algorithms, the observation that the Bayesian statistical approach achieves perfect inference of $N$ bits with less than $2 N$ one-bit measurements is still quite encouraging. We expect that the performance of the Bayesian active inference algorithms will be further improved after taking into account the possibility of ergodicity breaking in the version space of the perceptron. If the version space divides into a large number of well-separated clusters, the assumption of Gaussian distribution of the mean field theory will no longer be valid (Section~\ref{sec:discuss}), and more advanced mean field theories such as the fist-step replica-symmetry-breaking cavity method will be needed to better describe the complicated statistical correlations of the version space.

The concept of active (or adaptive) learning has been widely discussed in the fields of education science~~\cite{Chen-Li-etal-2018} and optimal experimental design~\cite{Box-1976,Robbins-1952}. Science itself may also be considered as an active learning process~\cite{Box-1976}, for which data-inspired intuitive insights, controlled experiments, and deductive reasoning are all indispensable. Artificial deep neural networks are becoming powerful tools for extracting the most important features from huge amount of data~\cite{LeCun-etal-2015,Goodfellow-Bengio-Courville-2016}, facilitating hypothesis formation and experimental design. Recently there has been great enthusiasm in this direction, and efficient active learning algorithms for deep neural networks are start to be explored~\cite{Huang-etal-2018, Rupprecht-Vural-2018,Ueltzhoeffer-2017,Friston-etal-2017}. A lot of work remains to be done on this important issue. From the theoretical side, the binary perceptron may serve as a simple model system to push for the limit of active Bayesian learning. Another basic inference problem which is closely related to the perceptron model is the so-called one-bit compressed sensing problem~\cite{Boufounos-Baraniuk-2008,Xu-Kabashima-2013}. At the moment only the passive mode of one-bit compressed sensing has been considered in the literature.  Beyond the single-layered perceptron and one-bit compressed sensing, the next and more challenging model is the multi-layered binary perceptron system.

\section{Inference by deductive reasoning}
\label{sec:logic}

\begin{figure}[b]
  \centering
  \includegraphics[angle=270,width=0.9\linewidth]{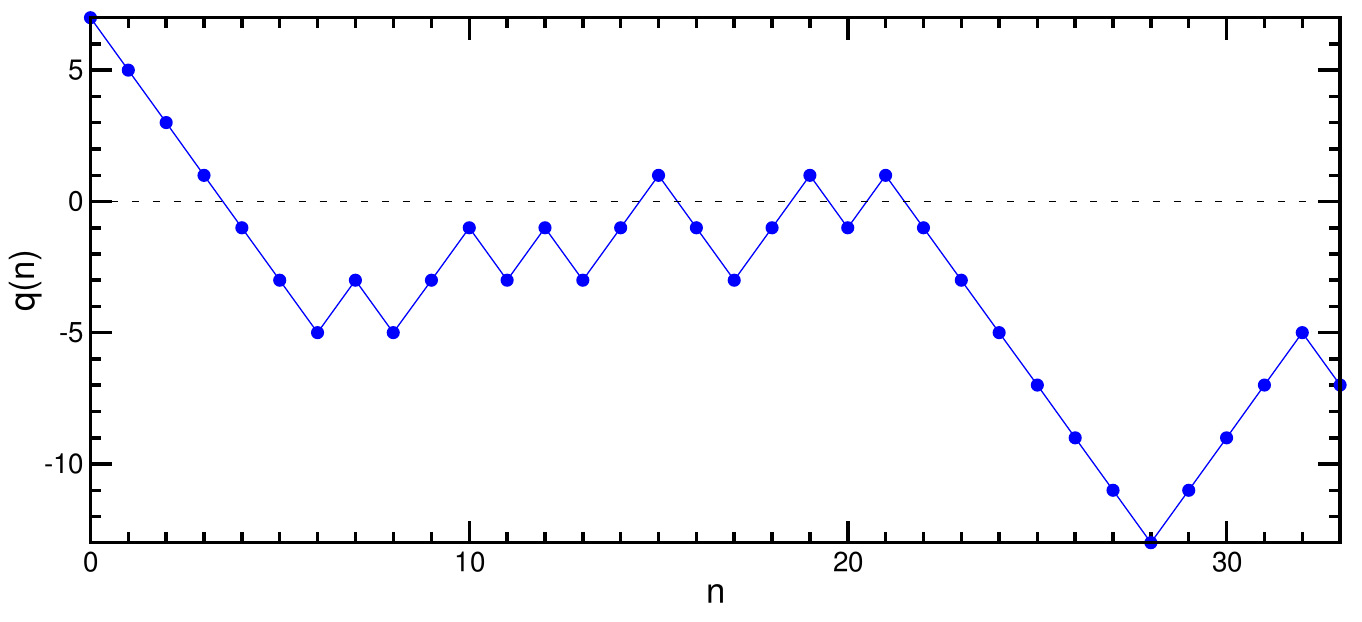}
  \caption{
    \label{fig:LogicN33}
    An example of the overlap function $q(n)$ for a small perceptron of size $N=33$. The random initial binary pattern $\bm{\xi}$ happens to have overlap $q(0)=7$ with the teacher's binary weight vector $\bm{T}$, and $q(n)$ is the new overlap with $\bm{T}$ after the first $n$ entries of $\bm{\xi}$ are all flipped. In this example $q(n)$ changes sign seven times.
  }
\end{figure}

We first show that if the student employs deductive reasoning, online perceptral learning can be made very efficient. At the start of the learning process, the student can simply choose an arbitrary pattern $\bm{\xi} = (\xi_1, \xi_2, \ldots, \xi_N) \in \{-1, +1\}^N$ as the query. For convenience of discussion, let us define a particular overlap function $q(n)$ on the integer domain $n\in\{0, 1, \ldots, N\}$ as
\begin{equation}
  \label{eq:qn}
  q(n) =
  \left\{
  \begin{array}{ll}
    \sum\limits_{i=1}^{N} \xi_i T_i & (n=0) \; , \\
    - \sum\limits_{i=1}^{n} \xi_i T_i + \sum\limits_{j=n+1}^{N} \xi_j T_j
    \quad \quad & (1 \leq n < N) \; , \\
    - \sum\limits_{i=1}^{N} \xi_i T_i & (n=N) \; .
  \end{array}
  \right.
\end{equation}
This function is simply the overlap (or the scalar product) of the teacher's weight vector $\bm{T}$ and the modified pattern $\bm{\xi}(n) \equiv (-\xi_1, \ldots, - \xi_n, \xi_{n+1}, \ldots, \xi_N)$ after flipping the first $n$ entries of $\bm{\xi}$. An example of function $q(n)$ is illustrated in Fig.~\ref{fig:LogicN33}. Because $N$ is odd, $q(n)$ takes only odd values. And because $q(N) = - q(0)$ and $| q(n) - q(n+1) | = 2$ for any $n<N$ (quasi-continuity), there must exist at least one $n^* \in \{0, 1, \ldots, N-1\}$ for which $|q(n^*)| = |q(n^*+1)|=1$ and $q(n^*+1) = -q(n^*)$.

The value of such an integer $n^*$ can be determined through at most $\log_2\!N$ queries. Starting from $n_{{\rm l}} = 0$ and $n_{{\rm r}}=N$, the query sequence goes as follows: (1) feed pattern $\bm{\xi}(n_{\rm l})$ to the perceptron to get the sign of $q(n_{{\rm l}})$;  (2) set $n_{{\rm m}} \equiv \lceil (n_{{\rm l}}+n_{{\rm r}})/2 \rceil$ and feed pattern $\bm{\xi}(n_{\rm m})$ to the perceptron to get the sign of $q(n_{{\rm m}})$; (3) if $q(n_{{\rm m}})$ and $q(n_{{\rm l}})$ have the same sign,  set $n_{{\rm l}} = n_{{\rm m}}$, otherwise set $n_{{\rm r}}=n_{{\rm m}}$; (4) if $n_{{\rm r}} = n_{{\rm l}} +1$, set $n^*=n_{{\rm l}}$ and quit, otherwise repeat steps (2)--(4).

The value of $q(n^*)$, i.e. the overlap between $\bm{\xi}(n^*)$ and $\bm{T}$, must be $+1$ or $-1$. Because of this fact, if one flips the $i$-th entry of $\bm{\xi}({n^*})$ and feeds the resulting slightly modified pattern to the perceptron, the value of $T_i$ can be deduced from the output. That is, if the output is the same as the sign of $q(n^*)$, then $T_i$ must be equal to $-\xi_i^* q(n^*)$, otherwise $T_i$ must be equal to $\xi_i^* q(n^*)$.

By the above-mentioned deductive reasoning method, the binary weight vector $\bm{T}$ can be exactly determined after at most $N+\log_2\!N$ queries. But it is clear from the above discussions that the success of this logical approach requires a deep understanding of the perceptron system. We continue to explore other active learning strategies in the next three sections.

\section{Version space minimization}
\label{sec:enumerate}

After $P$ training samples $(\bm{\xi}^\mu, \sigma^\mu)$ have been experienced, for a weight vector $\bm{J} = (J_1, J_2, \ldots, J_N) \in \{-1, +1\}^N$ to be compatible with all these $P$ samples, it must satisfy
\begin{equation}
  \label{eq:studentC}
        {\rm sign}\Bigl(\sum\limits_{i=1}^{N} J_i \xi_i^\mu \Bigr)  = \sigma^\mu
\end{equation}
for every one of these samples. We denote by $\Sigma_P$ the set of all the Ising weight vectors $\bm{J}$ satisfying these $P$ constraints, and refer to this set as the version space at stage $P$. Notice the teacher's vector $\bm{T}$ is always a member of $\Sigma_P$, so the volume $\bigl| \Sigma_P \bigr|$ of the version space must be positive-definite. To accelerate online learning, a simple idea would be to reduce the volume of the version space as much as possible with each new training pattern.  When finally the version space shrinks to a single point, this surviving element must be $\bm{T}$. Given the version space $\Sigma_P$ at stage $P$, then how should the student construct the next, $(P+1)$-th, Ising training pattern $\bm{\xi}^{P+1}$?

Consider two vectors $\bm{J}$, $\bm{J}^\prime$ of $\Sigma_P$. Each of them has equal probability to be the truth vector $\bm{T}$ (the uniform Bayesian prior distribution is assumed). If $\bm{J}$ happens to be the truth, then the other vector $\bm{J}^\prime$ can be refuted by a test pattern $\bm{\xi}$ if ${\rm sign}(\bm{\xi}; \bm{J}^\prime) = - {\rm sign}(\bm{\xi}; \bm{J})$. The probability of a randomly chosen vector $\bm{J}^\prime \in \Sigma_P$ being refuted by the test pattern $\bm{\xi}$ is then
\begin{equation}
  \frac{1}{2} \bigg[1- 
    \frac{{\rm sign}(\bm{\xi}; \bm{J})\sum_{\bm{J}^\prime \in \Sigma_P} 
      {\rm sign}(\bm{\xi}; \bm{J}^\prime)}{| \Sigma_P|} \biggr] \; .
\end{equation}
On the other hand, every $\bm{J} \in \Sigma_P$ is equally likely to be the truth $\bm{T}$, so we need to maximize the mean value of the above expression over all the different choices of $\bm{J}$,
\begin{equation}
  \frac{1}{2}\biggl[ 1 - \frac{\sum_{\bm{J}\in \Sigma_P}  
      {\rm sign}(\bm{\xi}; \bm{J}) }{|\Sigma_P|}
    \frac{\sum_{\bm{J}^\prime \in \Sigma_P}
      {\rm sign}(\bm{\xi}; \bm{J}^\prime)}{|\Sigma_P|} \biggr] \; ,
\end{equation}
which leads to the following constraint
\begin{equation}
  \label{eq:ALS}
  \sum\limits_{\bm{J} \in \Sigma_P} {\rm sign}(\bm{\xi}^{P+1}; \bm{J}) =  0 \; .
\end{equation}
In other words, $\bm{\xi}^{P+1}$ should be designed to divide the old version space $\Sigma_P$ into two parts of (almost) equal size. This designed pattern $\bm{\xi}^{P+1}$ must not be completely random, since the corresponding sum (\ref{eq:ALS}) for a completely random pattern will be of order $\sqrt{|\Sigma_P|}$. After $\bm{\xi}^{P+1}$ has been examined, one half of the members of $\Sigma_P$ will be discarded and the surviving vectors form the new version space $\Sigma_{P+1}$. Let us remark that this idea of version space bisection began to be discussed in the mathematics community in the early 1970s~\cite{Barzdin-Freivald-1972}, and it is underlying the widely appreciated optimal (minimum-error) Bayesian classification algorithm for the passive perceptron problem~\cite{Angluin-1988,Littlestone-1988,Opper-Haussler-1991}.

\begin{figure}[t]
  \centering
  \subfigure[]{
    \includegraphics[angle=270,width=1.0\linewidth]{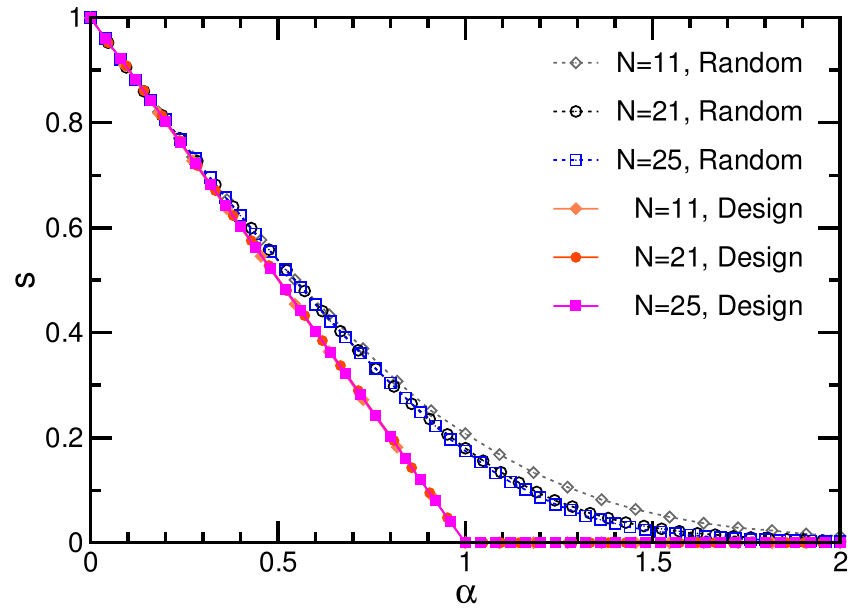}}
  \subfigure[]{
    \includegraphics[angle=270,width=1.0\linewidth]{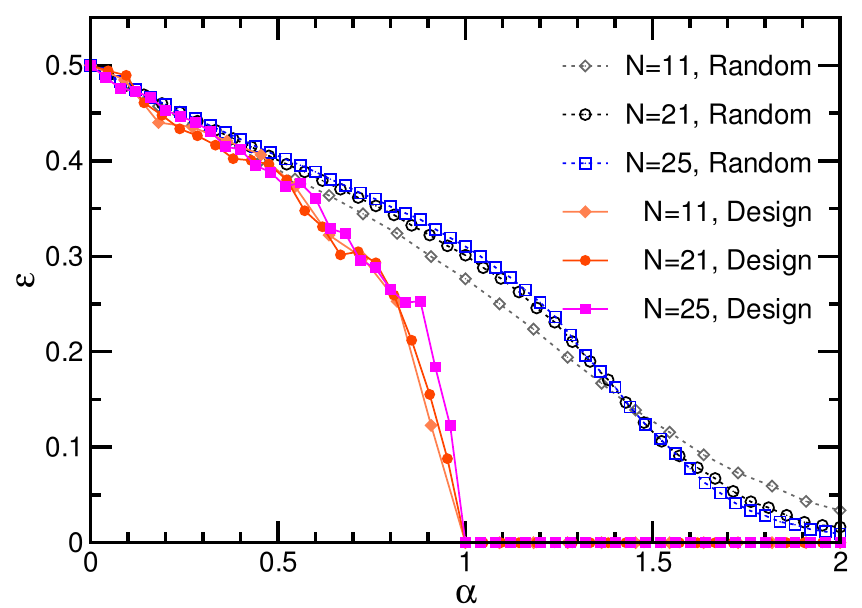}}
  \caption{
    \label{fig:IsingSmall}
    The active learning strategy (\ref{eq:ALS}) outperforms passive learning on small Ising perceptrons: (a) entropy density $s$ (in units of bit), (b) generalization error $\varepsilon$. Training patterns are added one after another, $\alpha = P/N$ is the instantaneous density of patterns ($N$ is the number of input neurons and $P$ is the number of training patterns). Each data point is obtained by averaging over $1,000$ independent runs of the passive (Random) or the active (Design) learning algorithm. }
\end{figure}

We test the performance of this conceptually simple designing principle on small perceptrons of size $N\leq 25$, for which the whole version space can be stored in the memory of a desktop computer.  In Figure~\ref{fig:IsingSmall} we show how the entropy density $s$ and the generalization error $\varepsilon$ change with the density $\alpha$ of training patterns. The entropy density measures the volume of the version space, $s \equiv \frac{1}{N} \log_2 |\Sigma_P|$. The generation error is computed as the probability that a randomly sampled binary test pattern will be mis-classified by a randomly chosen member $\bm{J}$ of the version space~\cite{Watkin-etal-1993,Engel-VanDenBroeck-2001}. If the training patterns are randomly and independently drawn from the configuration space $\{-1, +1\}^N$ (i.e., the passive learning mode), the mean entropy density $s$ and mean generalization error $\varepsilon$ both decrease gradually with $\alpha$ and are positive when $\alpha$ exceeds unity. On the other hand if the training patterns are required to satisfy (\ref{eq:ALS}) we find that the mean entropy density $s$ decreases linearly from $s=1$ to $s=0$ as $\alpha$ increases from $0$ to $1$, and at $\alpha=1$ the generalization error $\varepsilon$ becomes exactly zero.  In other words, perfect inference of the $N$-dimensional Ising vector $\bm{T}$ is achieved by the active strategy (\ref{eq:ALS}) with only $N$ one-bit queries (which only return $\pm 1$ values). Compared with the deductive reasoning approach, $\log_2\!N$ training patterns are saved following the designing principle (\ref{eq:ALS}). No algorithms can do better than this Bayesian strategy, because at least $N$ measurements are needed to exactly fix an $N$-dimensional vector.

These simulation results indicate that, in principle, perfect learning using only $N$ training patterns is possible. But directly employing Eq.~(\ref{eq:ALS}) to construct new training patterns is practically feasible only for small systems. When the dimension $N$ becomes large the version space $\Sigma_P$ (for $P$ small) will be too huge to enumerate. We must convert Eq.~(\ref{eq:ALS}) into a form suitable for implementation in large systems. This issue is addressed in the remaining part of this section.

With respect to all the accumulated $P$ training patterns at the end of the $P$-th learning stage, the volume of the version space $\Sigma_P$ (the partition function) is expressed as
\begin{equation}
  | \Sigma_P | = \sum\limits_{\bm{J}} \prod\limits_{\mu=1}^{P}
  \Theta\Bigl(\sigma^\mu \sum\limits_{j=1}^{N} J_j \xi_j^\mu \Bigr) \; ,
\end{equation}
where $\sigma^\mu$ is the true label of pattern $\bm{\xi}^\mu$, and $\Theta(x)$ is the Heaviside step function such that $\Theta(x)=1$ for $x> 0$ and $\Theta(x)=0$ for $x\leq 0$. The probability $\mathcal{P}_P(\bm{J})$ of weight vectors $\bm{J}$ in $\Sigma_P$ is
\begin{equation}
  \label{eq:pJjoint}
  \mathcal{P}_P(\bm{J}) = \frac{1}{| \Sigma_P |} \prod\limits_{\mu=1}^{P}
  \Theta\Bigl(\sigma^\mu \sum\limits_{j=1}^{N} J_j \xi_j^\mu \Bigr) \; .
\end{equation}
Notice that this probability distribution depends on the details of the $P$ training patterns. If $\bm{J} \in \Sigma_P$ we have $\mathcal{P}_P(\bm{J})=\frac{1}{|\Sigma_P|}$, otherwise $\mathcal{P}_P(\bm{J})=0$. Besides this joint distribution of all the $N$ entries $J_i$ of $\bm{J}$, we are also interested in single-weight marginals. The mean value of weight $J_i$ among all the vectors of $\Sigma_P$ is
\begin{equation}
  \label{eq:mJi}
  \langle J_i \rangle_P \equiv \frac{1}{|\Sigma_P|}
  \sum\limits_{{\bm J} \in \Sigma_P} J_i 
  = \sum\limits_{\bm{J}} \mathcal{P}_P(\bm{J}) J_i \; .
\end{equation}
Similarly, the mean value of $J_i J_j$ (pair correlation) is
\begin{equation}
  \langle J_i J_j \rangle_P \equiv \frac{1}{| \Sigma_P|}
  \sum\limits_{\bm{J}\in \Sigma_P} J_i J_j  = 
  \sum\limits_{\bm{J}} \mathcal{P}_P(\bm{J}) J_i J_j \; .
\end{equation}
When $i\!=\!j$ we have $\langle J_i J_i\rangle_P = 1$ due to the Ising nature of the weights. The training patterns bring correlations among the different weight variables. A consequence of these complicated correlations is that $\langle J_i J_j \rangle_P \neq \langle J_i \rangle_P \langle J_j \rangle_P$ for $i\!\neq\! j$.

Now consider adding a new training pattern $\bm{\xi}$ to the perceptron. The distribution $\mathcal{P}_P(q | \bm{\xi})$ of the overlap $q$ between $\bm{\xi}$ and the weight vectors $\bm{J}$ of $\Sigma_P$ is defined as
\begin{equation}
  \mathcal{P}_P(q | \bm{\xi})
  = \sum\limits_{\bm{J}} \mathcal{P}_P(\bm{J}) \delta\Bigl( q -
  \sum\limits_{i=1}^{N} \xi_i J_i \Bigr) \; ,
  \label{eq:qdist1}
\end{equation}
where $\delta(x)$ is the Dirac symbol such that $\delta(x)=1$ for $x=0$ and $\delta(x)=0$ for $x\neq 0$. From this definition we see that the mean overlap $q(\bm{\xi}) \equiv \int {\rm d} q \mathcal{P}_P(q | \bm{\xi}) q$ is
\begin{equation}
  q(\bm{\xi}) = \sum\limits_{i=1}^{N} \xi_i \langle J_i \rangle_P \; .
\end{equation}
The variance of the overlap $q$ is defined as $\Delta(\bm{\xi}) \equiv \int {\rm d} q \mathcal{P}_P(q | \bm{\xi}) q^2 - [q(\bm{\xi})]^2$. By a simple derivation we find that
\begin{equation}
  \Delta(\bm{\xi}) =
  \sum\limits_{i} \bigl(1 - \langle J_i \rangle_P ^2\bigr)
  + \sum\limits_{i< j} 2 \xi_i \xi_j
  \bigl( \langle J_i J_j \rangle_P - \langle J_i\rangle_P
  \langle J_j \rangle_P \bigr)  \; .
  \label{eq:qvariance}
\end{equation}
The overlap $q$ is the sum of $N$ random terms $\xi_i J_i$ (randomness coming from $J_i$). As the lowest-order approximation we assume that the central limit theorem is valid for $q$ when $N$ is large, even through the weights $J_i$ are not independent. In other words, we approximate the probability distribution (\ref{eq:qdist1}) by a Gaussian distribution:
\begin{equation}
  \label{eq:qdis}
  \mathcal{P}_P(q | \bm{\xi}) \approx \frac{1}{\sqrt{2 \pi \Delta(\bm{\xi})}}
  \exp\Bigl(-\frac{\bigl(q- q(\bm{\xi})\bigr)^2}
           {2 \Delta(\bm{\xi})}\Bigr) \; .
\end{equation}
(The possible breaking down of this Gaussian assumption will be discussed in Section~\ref{sec:discuss}.)

According to the designing principle (\ref{eq:ALS}), the $(P+1)$-th training pattern should refute half of the weight vectors in $\Sigma_{P}$. This means that the overlap between $\bm{\xi}^{P+1}$ and the weight vectors of $\Sigma_P$ should be positive for half of the elements $\bm{J} \in \Sigma_P$ and be negative for the remaining half. According to the Gaussian approximation (\ref{eq:qdis}), we see that the new pattern $\bm{\xi}^{P+1}$ should have zero mean overlap value with the weight vectors of $\Sigma_P$, that is
\begin{equation}
  \sum\limits_{i=1}^{N} \xi_i^{P+1} \langle J_i \rangle_P = 0 \; .
  \label{eq:APLmp}
\end{equation}
In comparison with Eq.~(\ref{eq:ALS}), the advantage of Eq.~(\ref{eq:APLmp}) is that the student does not need to memorize all the candidate truth vectors $\bm{J}$ but only need to evaluate the $N$ mean synaptic weights $\langle J_i \rangle_P$. Equation~(\ref{eq:APLmp}) may be regarded as a linearized version of Eq.~(\ref{eq:ALS}) with the highly nonlinear sign function replaced by a linear function. In the next section we discuss how to efficiently update the mean weights during the online learning process. We notice that the constraint (\ref{eq:APLmp}) is very similar in form to the designing constraint discussed in some of the early papers~\cite{Kinzel-Rujan-1990,Watkin-Rau-1992}. A significant difference is that our proposed constraint (\ref{eq:APLmp}) involves the mean synaptic weights $\langle J_i \rangle_P$, instead of the synaptic weights $J_i$ of a single weight vector stored in memory.

We employ simulated annealing~\cite{Kirkpatrick-etal-1983} to sample a maximally random pattern $\bm{\xi}^{P+1}$ under constraint (\ref{eq:APLmp}). An energy penalty is defined for each Ising pattern $\bm{\xi}$ as
\begin{equation}
  E(\bm{\xi}) = \Bigl| \sum_{i=1}^{N} \langle J_i \rangle_P \xi_i \Bigr| \; ,
\end{equation}
and the corresponding probability distribution of $\bm{\xi}$ is
\begin{equation}
  \label{eq:SAforXi}
  \mathcal{P}(\bm{\xi}) \propto \exp\bigl( - \beta E(\bm{\xi})\bigr)
  = \exp\Bigl( - \beta \Bigl|
  \sum_{i=1}^{N} \langle J_i \rangle_P \xi_i \Bigr| \Bigr) \; ,
\end{equation}
where the parameter $\beta$ is the inverse temperature. The pattern $\bm{\xi}$ evolves by single spin flips at slowly increasing $\beta$ values. At each elementary update step, (1) a randomly chosen entry $\xi_i$ of $\bm{\xi}$ is flipped to the opposite value; (2) if the energy difference $\Delta E \equiv E(\bm{\xi}^\prime) - E(\bm{\xi})$ between the modified pattern $\bm{\xi}^\prime$ and the old pattern $\bm{\xi}$ is non-positive, then this flip $\xi_i \rightarrow -\xi_i$ is surely accepted, otherwise it is accepted only with probability $e^{-\beta \Delta E}$. After $N$ such elementary flip trials the value of $\beta$ is then elevated by a constant factor $r_\beta$ ($\beta \leftarrow r_\beta \beta$). Finally, the spin configuration after a total number $t N$ of spin flip trials ($\beta$ has been recursively elevated $t$ times) is picked as the new training pattern $\bm{\xi}^{P+1}$. In this work we set $r_\beta = 1.1$ and $t=100$, and set the initial value of $\beta$ to be $0.01$. We have checked that the numerical results of the next two sections are insensitive to the particular values of these parameters.

\section{Experience accumulation}
\label{sec:hs1}

To exploit the designing principle (\ref{eq:APLmp}) we must first compute $\langle J_i \rangle_P$ for all the weight indices $i$. At $P=0$ we know that $\langle J_i \rangle_0 = 0$ for all the synaptic weights $J_i$. But the task for $P\geq 1$ is quite non-trivial and can not be made exact. In the online learning paradigm we compute $\langle J_i \rangle_P$ approximately by iteration.

After the training pattern $\bm{\xi}^{P+1}$ has been added to the system, the new distribution $\mathcal{P}_{P+1}(\bm{J})$ of the version space is related to the old $\mathcal{P}_{P}(\bm{J})$ through
\begin{equation}
  \mathcal{P}_{P+1}(\bm{J}) = 
  \frac{ \Theta\Bigl( \sigma^{P+1} \sum_{j} \xi_j^{P+1} J_j \Bigr) 
    \mathcal{P}_{P}(\bm{J})}{\sum_{\bm{J}^\prime}
    \Theta\Bigl(\sigma^{P+1} \sum_j \xi_j^{P+1} J_j^\prime \Bigr)
    \mathcal{P}_{P}(\bm{J}^\prime)} \; ,
\end{equation}
where $\sigma^{P+1} \equiv {\rm sign}\bigl(\sum_i T_i \xi_i^{P+1}\bigr)$ is the label of $\bm{\xi}^{P+1}$. The mean value of $J_i$ under $\mathcal{P}_{P+1}(\bm{J})$ is
\begin{equation}
  \label{eq:JiPp1}
  \langle J_i \rangle_{P+1} = \frac{p_i^+ A_i^+ - p_i^{-} A_i^-}
          {p_i^+ A_i^+ + p_i^- A_i^-} \; .
\end{equation}
Here, $p_i^+ \equiv (1+\langle J_i \rangle_P)/2$ is the probability of $J_i=+1$ in the version space $\Sigma_P$, and $p_i^{-}\equiv 1-p_i^{+}$ is the complementary probability; $A_i^+$ and $A_i^-$ are, respectively, the conditional probabilities of $\bm{\xi}^{P+1}$ being correctly classified by a weight vector $\bm{J}\in \Sigma_P$ given that $J_i = +1$ and $J_i = -1$. The defining expressions for $A_i^{+}$ and $A_i^{-}$ are 
\begin{subequations}
  \label{eq:Ai}
  \begin{align}
    A_i^{+} & \equiv \frac{1}{| \Sigma_P^{i+}|}
    \sum\limits_{\bm{J}\in \Sigma_P^{i+}} \Theta\bigl(\sigma^{P+1}\xi_i^{P+1}
    + \sigma^{P+1} \sum_{j\neq i} J_j \xi_j^{P+1}\bigr) \; ,
    \label{eq:Aip} \\
    A_i^{-} & \equiv \frac{1}{| \Sigma_P^{i-}|}
    \sum\limits_{\bm{J}\in \Sigma_P^{i-}} \Theta\bigl( - \sigma^{P+1}\xi_i^{P+1}
    + \sigma^{P+1} \sum_{j\neq i} J_j \xi_j^{P+1}\bigr) \; ,
    \label{eq:Aim}
  \end{align}
\end{subequations}
where $\Sigma_{P}^{i+}$ (respectively, $\Sigma_{P}^{i-}$) denotes the version sub-space of $\Sigma_P$ which contains all the vectors $\bm{J}\in \Sigma_P$ with $J_i=+1$ (respectively, $J_i=-1$). Notice that $\Sigma_{P}^{i+} \cap \Sigma_{P}^{i-} = \emptyset$ and $\Sigma_{P}^{i+} \cup \Sigma_{P}^{i-} = \Sigma_P$. Similar to the derivation of Eq.~(\ref{eq:qdis}) we can approximate $A_i^{+}$ and $A_i^{-}$ by two slightly different Gaussian integrals (details in Appendix~\ref{app:Jupdate}). Then we get the following convenient iterative equation for the mean weights
\begin{equation}
  \langle J_i \rangle_{P+1} \approx  \langle J_i \rangle_P
  + \sigma^{P+1} \xi_i^{P+1} \bigl(1-\langle J_i \rangle_P^2 \bigr) R_P \; .
  \label{eq:meanevol}
\end{equation}
Here $R_P$ is a magnitude factor determined by
\begin{equation}
  \label{eq:Rp}
  R_P \equiv \frac{2}{\sqrt{2\pi \Delta(\bm{\xi}^{P+1})}} 
  f\bigl( \frac{\sigma^{P+1} \sum_{k=1}^{N} 
    \xi_k^{P+1} \langle J_k \rangle_P}{\sqrt{2 \Delta(\bm{\xi}^{P+1})}}
  \bigr) \; ,
 \end{equation}
with the function $f(x)$ being
\begin{equation}
  \label{eq:IPfx}
  f(x) \equiv \frac{\exp(-x^2)}
  {1+\frac{2}{\sqrt{\pi}} \int_{0}^x \exp(-t^2) {\rm d} t} \; .
\end{equation}
The value of $f(x)$ equals to unity at $x=0$, and it then rapidly decays to zero as $x$ increases. When $x$ is negative $f(x)$ rapidly approaches the asymptotic form $-2 x$.

The iterative expression (\ref{eq:meanevol}) agrees with the belief-propagation equation reported in Ref.~\cite{Braunstein-Zecchina-2006} (see also Refs.~\cite{Kabashima-Uda-2004,Mezard-1989}). Notice that if $\xi_i^{P+1}$ has the same (respectively, opposite) sign of $\sigma^{P+1}$, the value of $\langle J_i \rangle_{P+1}$ increases (respectively, decreases) from $\langle J_i \rangle_P$ by an amount $(1- \langle J_i \rangle_P^2) R_P$. Equation (\ref{eq:meanevol}) therefore implements a specific Hebbian rule of experience accumulation. Notice also that $\langle J_i \rangle_P = \pm 1$ are two fixed points of Eq.~(\ref{eq:meanevol}), so if $\langle J_i \rangle_P$ is wrongly estimated to be (say) $-1$ while the truth value is $T_i= +1$, there is no chance to correct this mistake by further learning. To ensure that the iteration process is able to escape from a wrong fixed point, we therefore slightly modify Eq.~(\ref{eq:meanevol}) as follows:
\begin{equation}
  \langle J_i \rangle_{P+1} =  \langle J_i \rangle_P
  + \sigma^{P+1} \xi_i^{P+1} W\bigl(\langle J_i \rangle_P \bigr) R_P \; ,
  \label{eq:meanevolmod}
\end{equation}
where the weighting function $W\bigl(\langle J_i \rangle_P \bigr)$ by default is equal to $1-\langle J_i \rangle_P^2$, but $W\bigl(\langle J_i \rangle_P \bigr) = W_0$ if the sign of $\sigma^{P+1} \xi_i^{P+1}$ is opposite to that of $\langle J_i \rangle_P$ and at the same time $1-\langle J_i \rangle_P^2 < W_0$. The precise value of the cutoff parameter has a weak effect on the learning performance. After some preliminary experiments we finally set $W_0=8\times 10^{-3}$.

To determine the numerical value of the magnitude factor $R_P$, we need to compute the numerical value of the overlap variance $\Delta(\bm{\xi}^{P+1})$. The first summation of Eq.~(\ref{eq:qvariance}) is of order $N$ and is easy to compute. On the other hand, it is quite complicated to compute the second summation of Eq.~(\ref{eq:qvariance}). In this work we make the simplest approximation of independence among different weight variables, that is, $\langle J_i J_j \rangle_P \approx \langle J_i \rangle_P \langle J_j \rangle_P$. Under this additional approximation then
\begin{equation}
  \label{eq:qvarapp}
  \Delta(\bm{\xi}^{P+1}) \approx \sum\limits_{i=1}^{N}
  \bigl(1-\langle J_i \rangle_P^2 \bigr) \; ,
\end{equation}
and it is completely independent of $\bm{\xi}^{P+1}$. This independence approximate expression is commonly adopted in the literature (see, e.g., \cite{Braunstein-Zecchina-2006,Solla-Winther-1998}), it amounts to approximate the probability distribution $\mathcal{P}_P(\bm{J})$ by a factorized form. To improve the numerical accuracy of computing the mean weight values, the next step is to include at least partially the correlations between the weight elements (see, e.g., discussions in Refs.~\cite{Opper-1996,Solla-Winther-1998} concerning the continuous Perceptron). We leave this demanding issue for future work. (More discussion is made in Section~\ref{sec:discuss} on weight correlations.)

We now test the performance of the simple designing principle (\ref{eq:APLmp}). The following straightforward inference rule is adopted in the computer simulations: At the end of the $P$-th learning stage, the inferred truth vector $\hat{\bm{T}}^P \equiv (\hat{T}_1^P, \hat{T}_2^P, \ldots, \hat{T}_N^P)$ is simply the sign vector of the mean weights $\langle J_i \rangle_P$, with
\begin{equation}
  \hat{T}_i^P = {\rm sign}\bigl( \langle J_i \rangle_P \bigr) \; .
\end{equation}
The relative inference error is defined as the relative Hamming distance between the inferred weight vector $\hat{\bm{T}}^P$ and the teacher's weight vector $\bm{T}$, that is
\begin{equation}
  \label{eq:Hamming}
  d(\hat{\bm{T}}^P, \bm{T}) \equiv \frac{1}{N} \sum\limits_{i=1}^{N}
  \frac{|\hat{T}_i^P - T_i|}{2} \; .
\end{equation}
\begin{figure}[t]
  \centering
  \subfigure[]{
    \includegraphics[angle=270,width=1.0\linewidth]{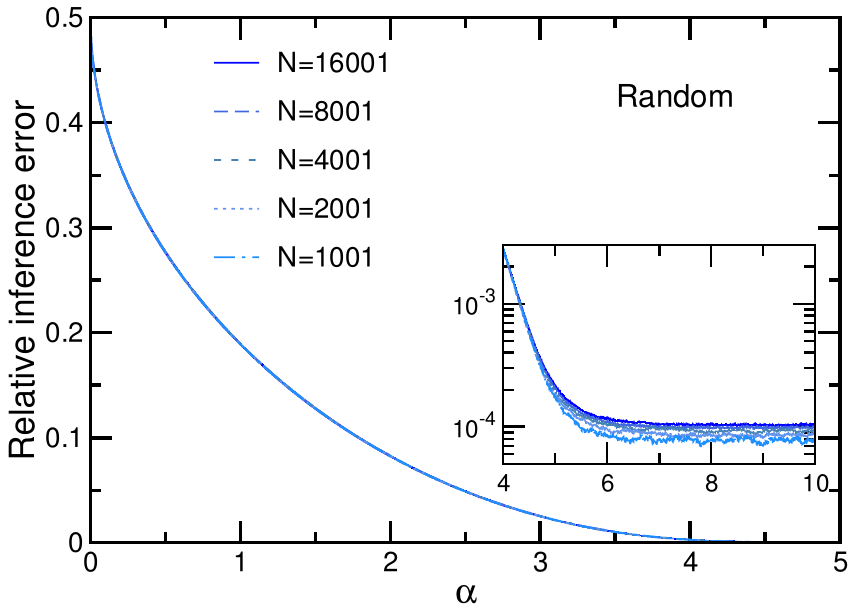}}
  \subfigure[]{
    \includegraphics[angle=270,width=1.0\linewidth]{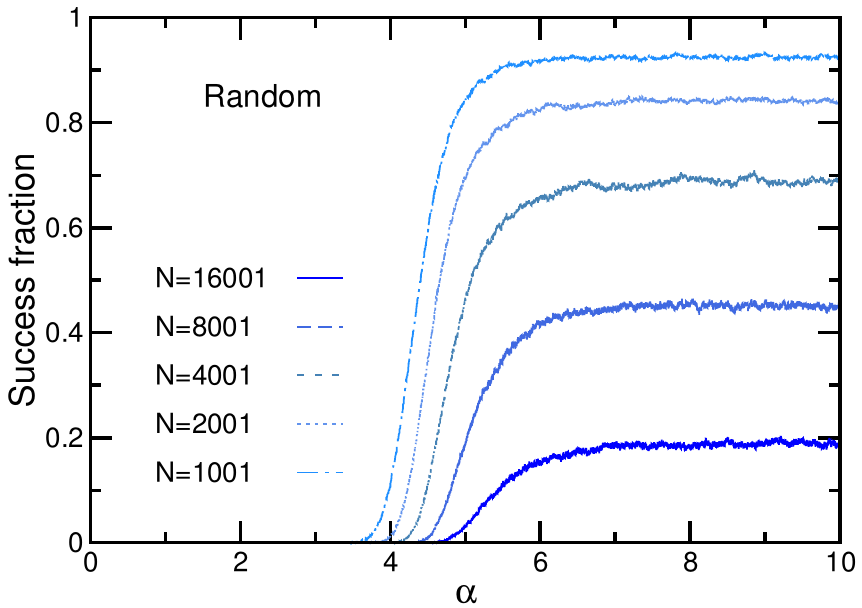}}
  \caption{
    \label{fig:ROIsing}
    The performance of passive online learning. The $P$ training patterns are fed to the student sequentially and they are independent random $N$-dimensional Ising vectors. The pattern density is $\alpha = P/N$. The total number of simulated independent online learning trajectories is $\mathcal{N}=10^4$. (a) The mean inference error, i.e., the mean fraction of incorrectly inferred teacher weights. The inset shows the tail part of the numerical data in semi-logarithmic scale. (b) The success fraction, i.e., the fraction of simulation trajectories in which the inferred weight vector is identical to the teacher's weight vector.
  }
\end{figure}
\begin{figure}[t]
  \centering
  \subfigure[]{
    \includegraphics[angle=270,width=1.0\linewidth]{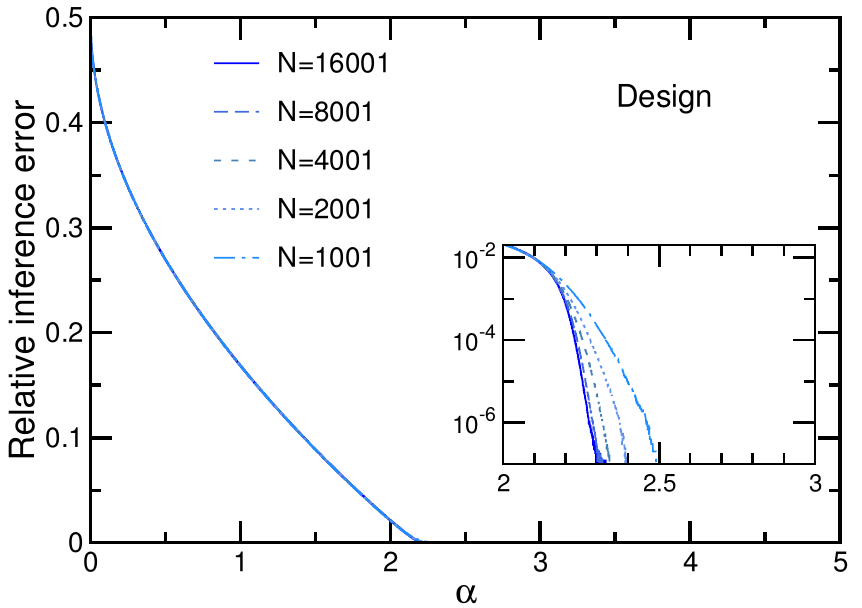}}
  \subfigure[]{
    \includegraphics[angle=270,width=1.0\linewidth]{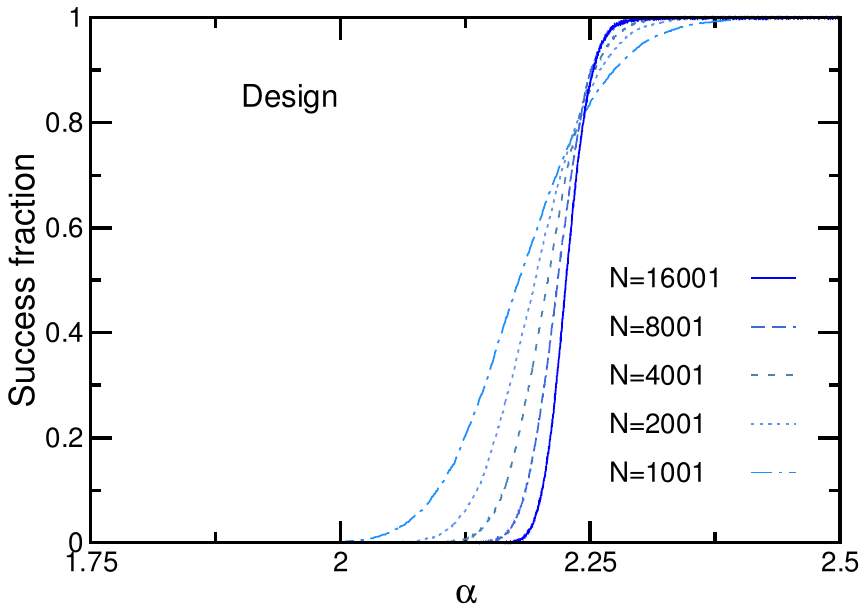}}
  \caption{
    \label{fig:DOIsing}
    Same as Fig.~\ref{fig:ROIsing}, but for active online learning under the designing principle (\ref{eq:APLmp}). Error-free inference is achieved at pattern density $\alpha\approx 2.23$.
  }
\end{figure}

Before testing the active learning mode, we first consider the random passive learning mode, in which every newly introduced training pattern is drawn independently and uniformly at random from the set of $2^N$ Ising patterns. The mean value of the relative inference error of this passive mode, averaged over $\mathcal{N}=10^4$ simulated independent online learning trajectories, is shown in Fig.~\ref{fig:ROIsing}(a) as a function of pattern density $\alpha$. We find that when $\alpha < 5$, the curves of mean inference error for different system sizes $N$ are well superimposed onto each other, while for $\alpha\geq 5$ the mean inference error saturates to a small positive level whose height slightly increases with system size $N$ [see inset of Fig.~\ref{fig:ROIsing}(a)].

As another measure of performance we consider the success fraction, which is defined as the probability that the inferred weight vector $\hat{\bm{T}}^P$ is identical to the truth vector $\bm{T}$ in repeated independent run of the whole online learning process. For example, if $\hat{\bm{T}}^P = \bm{T}$ in $4000$ out of $10,000$ independent runs then the success fraction is $0.4$ at this particular pattern density $\alpha=P/N$. Notice that the success fraction is not necessarily a monotonic function of $\alpha$ because it is possible that $\hat{\bm{T}}^{P} = \bm{T}$ but $\hat{\bm{T}}^{P+1} \neq \bm{T}$. The success fraction of passive online learning is shown in Fig.~\ref{fig:ROIsing}(b) as a function of pattern density $\alpha$. We find this fraction is first identical to zero for small $\alpha$ values, it then increases quickly when $\alpha$ exceeds $4$ and finally fluctuates around a plateau level. Since the height of the plateau level decreases considerably with system size $N$, error-free inference will be impossible in the thermodynamic limit of $N\rightarrow \infty$.

When constraint (\ref{eq:APLmp}) is imposed in designing each new training pattern, we find that the learning performance is greatly enhanced. As shown in Fig.~\ref{fig:DOIsing}(a), the mean inference error reaches below $10^{-3}$ as the pattern density $\alpha$ is increased up to $2.2$ and it then quickly drops to zero as $\alpha$ further increases slightly. The dramatic effect of active learning is most clearly demonstrated by the difference between Fig.~\ref{fig:DOIsing}(b) and Fig.~\ref{fig:ROIsing}(b). We see from Fig.~\ref{fig:DOIsing}(b) that the curve of success fraction becomes more and more sharper as the system size $N$ increases, and all these different curves intersect at approximately the same value of $\alpha$. Similar system size-dependent behaviors are commonly observed in finite-size scaling studies of continuous phase transitions~\cite{Privman-1990}. We conjecture that a well-defined dynamical phase-transition to perfect inference will occur at the value of $\alpha\approx 2.23$ in the thermodynamic limit of $N\rightarrow \infty$. More thorough theoretical investigation on the large $N$ limit of this learning dynamics will be carried in a follow-up paper.

\section{Additional orthogonality considerations}
\label{sec:hs2}

When a new training pattern $\bm{\xi}^{P+1}$, constrained by Eq.~(\ref{eq:APLmp}) but otherwise being maximally random, is sampled by simulated annealing [Eq.~(\ref{eq:SAforXi})], it is conditionally independent of all the earlier training patterns given the values of $\langle J_j \rangle_P$. But since the mean weights $\langle J_j \rangle_P$ are determined through the accumulative mechanism (\ref{eq:meanevolmod}), Eq.~(\ref{eq:APLmp}) indeed brings complicated correlations between $\bm{\xi}^{P+1}$ and all its predecessors. If $\bm{\xi}^{P+1}$ happen to be relatively similar to some of the old training patterns, its power in promoting active inference will be compromised~\cite{Kinouchi-Caticha-1992,Shinzato-Kabashima-2009}. According to the geometric picture underlying the exact designing principle (\ref{eq:ALS}), it should be beneficial to explicitly require (at least approximate) orthogonality between $\bm{\xi}^{P+1}$ and the training patterns introduced during the last $M$ steps.

To implement these additional orthogonality constraints, we modify the energy function of the simulated annealing process as follows:
\begin{equation}
  \label{eq:Enew}
  E(\bm{\xi}) = \Bigl| \sum\limits_{i=1}^{N} \langle J_i \rangle_P \xi_i  \Bigr|
  + \lambda \frac{\sum\limits_{\mu=1}^{\min(P, M)}
    \bigl(\sum\limits_{j=1}^N \xi_j^{P+1-\mu} \xi_j \bigr)^2 }{\min(M, P)} \; .
\end{equation}
The second energy term is equal to the average squared overlap between $\bm{\xi}$ and an old pattern $\bm{\xi}^\mu$. The parameter $\lambda$ controls the relative importance of the additional orthogonality constraints. In the present work we set $\lambda = 1$, and considering that there are $N$ mutually orthogonal vectors in an $N$-dimensional space, we set $M=N-1$. (We have not yet tried to optimize the values of $\lambda$ and $M$.) Each new training pattern is then sampled by simulated annealing starting from an initial completely random pattern. The only difference is that the energy function in Eq.~(\ref{eq:SAforXi}) now takes the form of Eq.~(\ref{eq:Enew}).

\begin{figure}[t]
  \centering
  \subfigure[]{
    \includegraphics[angle=270,width=1.0\linewidth]{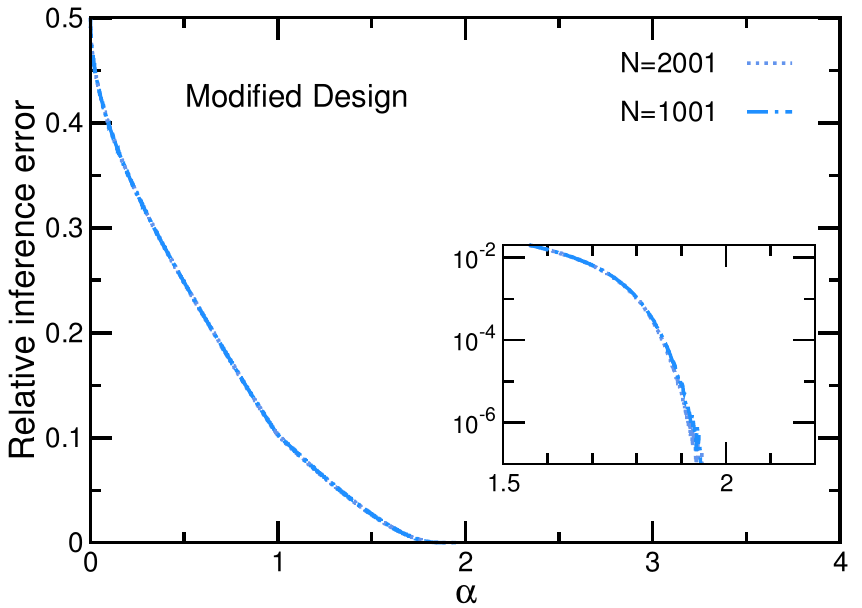}}
  \subfigure[]{
    \includegraphics[angle=270,width=1.0\linewidth]{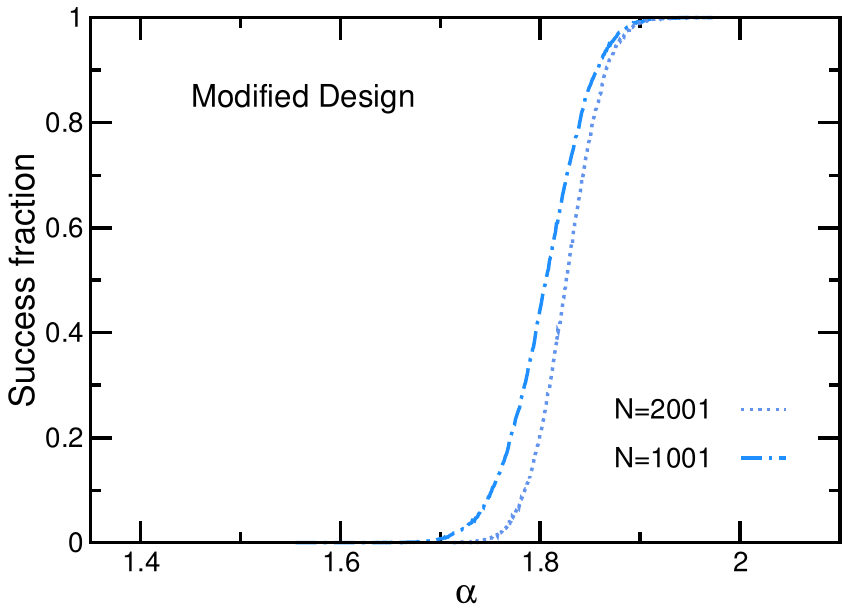}}
  \caption{
    \label{fig:DOMIsing}
     Same as Fig.~\ref{fig:ROIsing} and Fig.~\ref{fig:DOIsing}, but for active online learning under the designing constraint (\ref{eq:APLmp}) and the additional orthogonality constraints, see Eq.~(\ref{eq:Enew}). The number of stored training patterns is set to be $M=N\!-\!1$.
  }
\end{figure}

The performance of the modified online learning algorithm is shown in Fig.~\ref{fig:DOMIsing}. We find that error-free inference of the teacher's weight vector can be achieved with high probability after encountering $P\!\approx\!1.9 N$ training patterns. Compared with the results of Fig.~\ref{fig:DOIsing}, the additional orthogonality considerations indeed lead to a remarkable boost to the learning efficiency. 

It may be possible to further improve the learning performance by optimizing the parameters $\lambda$ and $M$ of Eq.~(\ref{eq:Enew}). Furthermore, the energy function (\ref{eq:Enew}) may not necessarily be the best way to incorporate both the designing constraint (\ref{eq:APLmp}) and the additional orthogonality constraints. For example, it may be even better to consider all the $P$ old patterns (instead of only the last $M$ ones) with non-uniform weighting factors.

\section{Discussion}
\label{sec:discuss}

In this work we considered the Bayesian active learning principle (\ref{eq:ALS}) to infer the teacher's weight vector of an $N$-dimensional Ising perceptron. Each new Ising training pattern is not randomly drawn as in passive learning but is designed with the aim of splitting the current version space into two equal sub-spaces. This designing principle was exactly implemented for small systems to achieve error-free inference using only $N$ training samples (Fig.~\ref{fig:IsingSmall}). When exhaustive enumeration becomes unfeasible for large systems, we derived a simple constraint (\ref{eq:APLmp}) based on this principle and demonstrated that error-free inference is achievable with $P\!\approx\!2.3 N$ training samples (Fig.~\ref{fig:DOIsing}). The number of training samples was further reduced to $P\!\approx\!1.9 N$ after imposing additional orthogonality constraints on the training patterns (Fig.~\ref{fig:DOMIsing}). On the other hand, the deductive reasoning algorithm discussed in this paper is guaranteed to achieve error-free inference with at most $N\!+\!\log_2\!N$ queries and therefore is much superior to the Bayesian strategies discussed in Secs~\ref{sec:hs1} and \ref{sec:hs2}.

In deriving the constraint Eq.~(\ref{eq:APLmp}) from the Bayesian principle (\ref{eq:ALS}), we have approximated the overlap probability profile $\mathcal{P}_P(q | \bm{\xi})$ [Eq.~(\ref{eq:qdist1})] by a Gaussian distribution. Maybe this Gaussian assumption is only valid for sufficiently small values of the pattern density $\alpha$. With the addition of training patterns, the volume of the version space becomes more or more small. At the same time the shape of the version space may become more and more irregular. One highly possible scenario is that, when $\alpha$ exceeds certain threshold value, the version space breaks up into many well-separated sub-spaces with each of them having a different set of mean weight values $\{\langle J_i \rangle_P\}$. As a consequence, the overlap probability profile $\mathcal{P}_P(q | \bm{\xi})$ should be described as a weighted sum of many distinct Gaussian distribution functions (one for each version sub-space), and then Eq.~(\ref{eq:APLmp}) should be modified accordingly.

From the academic point of view, active learning in the presence of ergodicity breaking is a very interesting challenge. With an accurate approximation to the overlap probability profile $\mathcal{P}_P(q | \bm{\xi})$, the efficiency of the active learning process may closely approach the theoretical limiting value of $\alpha\!=\!1$. We hope that significant theoretical and algorithmic progresses will be made in the near future on this important research issue.

\section*{Acknowledgement}

This work was supported by the National Natural Science Foundation of China (grant numbers 11421063 and 11747601) and the Chinese Academy of Sciences (grant number QYZDJ-SSW-SYS018). Numerical simulations were ccarrried out at the HPC cluster and Tianwen cluster of ITP-CAS. The author thanks Kim Sneppen for valuable discussions.

\begin{appendix}

\section{Derivation of Eq.~(\ref{eq:meanevol})}
\label{app:Jupdate}

The quantity $A_i^{+}$ as defined by Eq.~(\ref{eq:Aip}) is just the conditional probability of the random value $x$ defined by $x = \sigma^{P+1}( \xi_i^{P+1} + \sum_{j\neq i} \xi_j^{P+1} J_j)$ being positive among all the weight vectors $\bm{J} \in \Sigma_P^{i+}$. The mean value of this random variable $x$ is
\begin{equation}
  \sigma^{P+1} \Bigl( \xi_i^{P+1} + \sum_{j\neq i} \xi_j^{P+1}
  \langle J_j \rangle_P^{i+} \Bigr) \; ,
\end{equation}
and its variance is
\begin{eqnarray}
 & & \hspace*{-0.5cm}
  \sum\limits_{j\neq i} \bigl(1 - (\langle J_j \rangle_{P}^{i+})^2 \bigr) +
  \nonumber \\
  & & \sum\limits_{j < k}^\prime 2 \xi_j^{P+1} \xi_k^{P+1}
    \bigl( \langle J_j J_k \rangle_{P}^{i+} - \langle J_j \rangle_{P}^{i+} 
  \langle J_k \rangle_{P}^{i+} \bigr) \; ,
  \label{eq:qvarnoti}
\end{eqnarray}
where $\sum_{j< k}^\prime$ means the summation is over pair of indices $j, k \in \{1, 2, \ldots, N \}$ satisfying $j<k$ and $j, k \neq i$. In the above two expressions, $\langle J_j \rangle_{P}^{i+}$ denotes the mean value of $J_j$ in the version space $\Sigma_{P}^{i+}$ and similarly for $\langle J_j J_k \rangle_{P}^{i+}$.

To proceed, let us approximate $\langle J_j \rangle_{P}^{i+}$ simply by $\langle J_j \rangle_P$ and assume that $\langle J_j J_k \rangle_{P}^{i+} = \langle J_j \rangle_{P}^{i+} \langle J_k \rangle_{P}^{i+}$. The variance (\ref{eq:qvarnoti}) then simplifies to be $\Delta(\bm{\xi}^{P+1}) - (1- \langle J_i \rangle_P^2)$ and after neglecting the correction term $(1-\langle J_i \rangle_P^2)$, is finally approximated to be $\Delta(\bm{\xi}^{P+1})$, where $\Delta(\bm{\xi}^{P+1})$ is computed according to Eq.~(\ref{eq:qvarapp}). Under the assumption that the probability profile of the random variable $x$ is well approximated by a Gaussian distribution, we obtain the following integration expression for the conditional probability $A_i^{+}$:
\widetext
\begin{equation}
  \label{eq:Aip2}
  A_{i}^+ = \frac{1}{\sqrt{2 \pi \Delta(\bm{\xi}^{P+1})}} \int\limits_{0}^{+\infty}
  \exp\Bigl( - \frac{ \bigl( x - \sigma^{P+1} \xi_i^{P+1} - \sigma^{P+1}
    \sum_{j\neq i} \xi_j^{P+1} \langle J_j \rangle_P \bigr)^2}
           {2 \Delta(\bm{\xi}^{P+1})} \Bigr)  {\rm d} x \; .
\end{equation}
An approximate expression for the conditional probability $A_i^{-}$ as defined by Eq.~(\ref{eq:Aim}) can be derived by the same way:
\begin{equation}
  \label{eq:Aim2}
  A_{i}^- = \frac{1}{\sqrt{2 \pi \Delta(\bm{\xi}^{P+1})}} \int\limits_{0}^{+\infty}
  \exp\Bigl( - \frac{ \bigl( x + \sigma^{P+1} \xi_i^{P+1} - \sigma^{P+1}
    \sum_{j\neq i} \xi_j^{P+1} \langle J_j \rangle_P \bigr)^2}
           {2 \Delta(\bm{\xi}^{P+1})} \Bigr)  {\rm d} x \;. 
\end{equation}

The above two expressions are still not very convenient for numerical computations. By treating $\xi_i^{P+1} (1-\langle J_i \rangle_P)$ and $\xi_i^{P+1}(1 + \langle J_i \rangle_P)$ as expansion small quantities with respect to the sum $\sum_{j=1}^{N} \xi_j^{P+1} \langle J_j \rangle_P$, we obtain that
\begin{subequations}
  \label{eq:Aiapp3}
  \begin{align}
    A_i^{+} & = A + \xi_i^{P+1} (1-\langle J_i \rangle_P) \delta A \; , \\
    A_i^{-} & = A - \xi_i^{P+1} (1- \langle J_i \rangle_P) \delta A \; ,
  \end{align}
\end{subequations}
where
\begin{subequations}
  \begin{align}
    A & = \frac{1}{\sqrt{2 \pi \Delta(\bm{\xi}^{P+1})}} \int\limits_{0}^{+\infty}
    \exp\Bigl(- \frac{ \bigl(x- \sigma^{P+1} \sum_{j=1}^N \xi_j^{P+1} \langle J_j
      \rangle_P \bigr)^2}{2 \Delta(\bm{\xi}^{P+1})} \Bigr) {\rm d} x \; ,
    \\
    \delta A  & = \frac{1}{\sqrt{2\pi \Delta(\bm{\xi}^{P+1})}}
           \exp\Bigl(-\frac{\bigl( \sum_{j=1}^{N} \xi_j^{P+1} \langle J_j
             \rangle_P\bigr)^2}{2 \Delta(\bm{\xi}^{P+1})} \Bigr) \; .
  \end{align}
\end{subequations}
By inserting Eq.~(\ref{eq:Aiapp3}) into Eq.~(\ref{eq:JiPp1}) we obtain the experience accumulation formula (\ref{eq:meanevol}).

\vskip 0.5cm

\endwidetext

\end{appendix}


\end{document}